\documentclass[aps,pra,reprint,longbibliography]{revtex4-1}

\usepackage{graphicx}
\usepackage{amsmath}
\usepackage{amssymb}
\usepackage{textcomp}

\begin{document}
	
\title[]{Superconducting Optoelectronic Neurons \uppercase\expandafter{\romannumeral 1 \relax}: General Principles}
	
\author{Jeffrey M. Shainline, Sonia M. Buckley, Adam N. McCaughan, Jeff Chiles, Richard P. Mirin, and Sae Woo Nam}
\affiliation{National Institute of Standards and Technology, 325 Broadway, Boulder, CO, 80305}

\date{\today}
	
\begin{abstract}
The design of neural hardware is informed by the prominence of differentiated processing and information integration in cognitive systems. The central role of communication leads to the principal assumption of the hardware platform: signals between neurons should be optical to enable fanout and communication with minimal delay. The requirement of energy efficiency leads to the utilization of superconducting detectors to receive single-photon signals. We discuss the potential of superconducting optoelectronic hardware to achieve the spatial and temporal information integration advantageous for cognitive processing, and we consider physical scaling limits based on light-speed communication. We introduce the superconducting optoelectronic neurons and networks that are the subject of the subsequent papers in this series.
\end{abstract}
	
	
\maketitle
	
\section{\label{sec:introduction}Introduction}
Complete understanding of the information processing underlying cognition remains a significant scientific challenge. Progress in neuroscience, computer science, psychology, and neural engineering make this a fruitful time for elucidation of intelligence. Biological experiments and software simulations would be greatly augmented by artificial hardware with complexity comparable to systems we know to be conscious. Intelligent systems implemented with hardware optimized for neural computing may inform us regarding the limits of cognition imposed by the speed of light while providing technological opportunities sufficient to spawn a new domain of the computing industry.

As we will argue, neural computing appears uniquely capable of the distributed, yet integrated, information processing that characterizes intelligent systems. Many approaches to neural computing are being developed, and the maturity of the semicondutor industry makes CMOS a wise initial platform. Yet the central role of communication in neural computing indicates that hardware incorporating different physics may be advantageous for this application. In previous work \cite{shbu2017}, we considered the potential for superconducting optoelectronic hardware to perform neural operations. The principal assumption guiding the design of the hardware platform is that photons are the entities best suited for communication in large-scale neural systems. The hardware platform leverages optical communication over short and long distances to enable dense local fanout as well as distant communication with the shortest possible delay. In a series of papers \cite{sh2018b,sh2018c,sh2018d,sh2018e}, we present details of the design of superconducting optoelectronic neurons and networks that appear capable of achieving the functions required for cognitive computing. In this paper, we summarize the physical reasoning behind the hardware to meet the requirements of cognitive circuits and provide an overview of the operation of the neurons. 

The audience we hope to address is broad and includes neuromorphic engineers, perhaps studied in silicon but open to new exploration; the integrated-photonics community, who may see this as a promising application of photonic devices and systems; the superconducting electronics community, who may find benefits to long-standing challenges such as memory, clock distribution, cryogenic I/O, and achieving the voltage necessary to interface with CMOS; neuroscientists, who may utilize this hardware platform to test hypotheses at device and system levels; and the advanced computing community, who may leverage the capabilites of these systems to solve outstanding problems.

\section{\label{sec:cognitiveSystems}Cognitive systems}
The foundational assumption of this work is that light is the physical entity best-suited to achieve communication in cognitive neural systems. To motivate why light is essential for large-scale neural systems, we must describe the systems we intend to pursue.

Broadly speaking, we wish to pursue devices and networks capable of acquiring and assimilating information across a wide range of spatial, temporal, and categorical scales. In a neural cognitive system, spatial location within the network may correspond to information specific to content area or sensory modality, and therefore spatial integration across the network corresponds to integration across informational subjects and types. Information processing must occur across many levels of hierarchy with effective communication across local, regional, and global spatial scales as well as temporal scales. These systems must continually place new information in context. It is required that a cognitive system maintain a slowly varying background representation of the world while transitioning between dynamical states under the influence of stimulus. The objective of this series of papers is to design general cognitive circuits with structural and dynamical attributes informed by neuroscience, network theory, and dynamical systems. Stated generally, systems combining functional specialization with functional integration are likely to perform well for many cognitive tasks \cite{spto2000,spto2002}.

The theme of localized, differentiated processing combined with information integration \cite{toed1998,tosp2003,to2004,seiz2006,bato2008,bato2009,base2011} across space \cite{busp2009,sp2010} and time \cite{sase2001,vala2001,enfr2001,budr2004,bu2006} is central to the device and network designs we present here. In the spatial domain, the demand for integration of information from many local areas requires dense local connectivity (as measured by a clustering coefficient \cite{eskn2015,saki2007,fa2007}), but also connections between these local areas which serve to combine the local information and place it in a larger context at higher cognitive levels \cite{brto2006} (as measured by a short average path length \cite{alba2002}). High clustering combined with short average path length defines a small-world network \cite{wast1998}. For the highest performance, we expect this trend of integration of locally differentiated information to repeat across many scales in a nearly continuous manner \cite{busp2009,sp2010} such that any node in the system is likely to be processing information with local neighbors, but also receiving information from simpler, less-connected units, and transferring information to complex, highly connected units. Networks with this organization across scales are governed by power law spatial scaling \cite{baal1999}.

The patterns are related in the temporal domain where transient synchronized oscillations integrate information from various brain regions \cite{vala2001,sase2001,enfr2001}. Information exchange can occur on very fast time scales, and results of these computations must be combined over longer times. The spatial structure of the network and its operation in the time domain are not independent \cite{spto2000,spto2002,waxu2006}. Fast, local dynamics integrate information of closely related nodes through transient neuronal functional clusters \cite{buwa2012}, while activity on slower scales can incorporate input from larger regions \cite{stsa2000}. Networks with this organization in time are governed by a power law frequency distribution \cite{bata1987,budr2004,bu2006}, characteristic of self-organized criticality \cite{be2007}. Power law spatial and temporal distributions underlie systems with fractal properties \cite{bata1987,bu2006}, and self-similarity across space and time is advantageous for cognition \cite{bu2006,be2007,kism2009,shya2009,ch2010,rusp2011}. 

These conceptual arguments regarding information integration across spatial and temporal scales lead us to anticipate networks with hierarchical configuration, with processing on various scales being integrated at high levels to form a coherent cognitive state \cite{brto2006}. The constitutive devices most capable of achieving these network functions are relaxation oscillators \cite{bu2006,st2015}, dynamical entities characterized by pulsing behavior \cite{mist1990} with resonant properties at many frequencies \cite{soko1993,huya2000}. Neurons are a subset of relaxation oscillators with complex operations adapted for spike-based computation \cite{geki2002}.

To illustrate how differentiated processing and information integration are implemented by neurons for cognition, consider vision \cite{laus2011}. In early stages of visual processing, neurons located near each other in space will show similar tuning curves \cite{daab2001} in response to presented stimuli, thus forming locally coherent assemblies selecting for certain features of a visual scene \cite{enfr2001}. These locally differentiated processing units are constructed from architectural motifs \cite{spko2004,onsa2005} and are manifest in biological hardware as mini-columns and columns \cite{mo1997}, which are dedicated to modeling a subset of sensory space \cite{haah2017}. To form a more complete representation of an object within a visual scene, or to make sense of a complex visual scene with many objects, the visual system must combine the information from many differentiated processors. This integration is accomplished with lateral connections between columns \cite{spto2000} as well as with feed-forward connections from earlier areas of visual cortex to later areas of visual cortex \cite{laus2011}. Such an architecture requires some of the neurons in any local region to have long-range projections, motivating the need for local connectivity for differentiated processing combined with distant connectivity for information integration across space. 

Temporal considerations are as important as spatial, yet more subtle. To understand information integration in the time domain, consider synchronized oscillations at various frequencies in the context of the binding problem \cite{ro1999,tr1999}. Stated as a question, the binding problem asks how the myriad stimuli presented to the brain can be quickly and continuously organized into a coherent cognitive moment. In the limited context of vision, we ask how a complex, dynamic visual scene can be structured into a discernible collection of objects that can be differentiated from each other and from an irrelevant background \cite{rede1999}. Many studies provide evidence that fast, local oscillations are modulated by slower oscillations encompassing more neurons across a larger portion of the network \cite{vala2001,sase2001,enfr2001,lued1997,stsa2000,budr2004,bu2006,fr2015}. In the case of columns in visual cortex, local clusters tuned to specific stimuli will form assemblies with transient synchronization at high frequencies ($\gamma$ band, 20-80\,Hz \cite{budr2004}). The information from many of these differentiated processors is integrated at higher levels of processing by synchronizing larger regions of neurons at lower frequencies ($\alpha$ band, 1-5\,Hz, and $\theta$ band, 4-10\,Hz \cite{stsa2000,budr2004}). The transient synchronization of neuronal assemblies is closely related to neuronal avalanches \cite{be2007,shya2009}, cascades of activity across all these frequencies. Neuronal avalanches are observed in networks balanced at the critical point between order and chaos \cite{be2007,kism2009,shya2009,ch2010,rusp2011}. Self-similarity in the temporal domain implies operation at this critical point \cite{be2007,kism2009,rusp2011}, and operating at this phase transition is necessary to maximize the dynamic range of the network \cite{shya2009}. Inhibition and activity-based plasticity are crucial for achieving this balance \cite{budr2004,bu2006,siqu2007}.  

Networks of excitatory principal neurons interspersed with inhibitory interneurons \cite{robu2015} with small-world characteristics naturally synchronize at frequencies determined by the circuit and network properties \cite{bu2006}. Slower frequency collective oscillations of networks of inhibitory interneurons provide short windows when certain clusters of excitatory neurons are uninhibited and therefore susceptible to spiking \cite{buge2004}. This feedback through the inhibitory interneuron network provides a top-down means by which the dynamical state of the system can provide broad information to the local processing clusters \cite{enfr2001,fr2015}. Regions of cortex with higher information integration focus attention \cite{vala2001} on certain aspects of stimulus by opening receptive frequency windows at the resonant frequencies of relevant sub-processors, providing a mechanism by which binding occurs and background is ignored \cite{lued1997,enfr2001,budr2004,fr2015}. The result of this inhibitory structuring of time is a network with dynamic effective connectivity \cite{brto2006,fr2015}. By constructing a network with small-world, power-law architecture from highly tunable relaxation oscillators, and employing feedback through inhibitory oscillations, we produce a system that can change its effective structural and resonant properties very rapidly based on information gleaned from prior experiences of a large region of the network \cite{budr2004,fr2015}.

This model of binding requires a means by which the resonant frequencies of neuronal assemblies can be associated with certain stimuli, and a means by which the inhibitory interneuron network can learn to associate different assemblies with different frequencies. Plastic synaptic weights make such adaptation possible. Synapses provide a means by which the connectivity of the network can shape dynamics and functionality, and synapses adapt their states based on internal and external activity. As cortex evolves through dynamical states on various temporal and spatial scales, information stored in synapses is integrated. This dynamical state integrates synaptic information across the network, and uses this information as feedback to distributed sub-processors \cite{enfr2001,fr2015}. 

For a cognitive system embedded in a dynamical environment to provide adaptive feedback as well as robust memory, the system must comprise a large number of synapses changing on different time scales due to different internal and external factors \cite{fudr2005}. Synapses with many stable values of efficacy can significantly increase memory retention times \cite{fuab2007}, and synapses that adapt not only their state of efficacy but also their probability of state transition are crucial for maximizing memory retention times \cite{fudr2005,khso2017}. Adaptation of probability of state transition is a mechanism of metaplasticity \cite{ab2008}, and many forms appear in biological systems, which employ many techniques for extending memory retention \cite{ab2008}. We expect a cognitive system to utilize differentiated regions of neurons, some with synapses changing readily between only two synaptic states, and other regions with synapses changing slowly between many distinguishable states. We further expect the network to update not only synaptic weights but also the probability of changing synaptic weights. The dynamical state of the system can then sample synaptic memory acquired at many times, in many contexts, while quickly adapting the dynamical trajectory as new stimulus is presented.

To summarize, cognition appears to require differentiated local processing combined with information integration across space, time, and experience. The structure of the network determines the dynamical state space, and the structure of the network adapts in response to stimulus and internal activity. We now ask the question: what physical systems are best equipped to perform these operations?
	
\section{\label{sec:physicsAndHardware}Physics and hardware for cognition}
The aforementioned insights from neuroscience lead us to emphasize several features of neural systems in hardware for cognition. First, we must use a physical signaling mechanism capable of achieving communication across networks with dense local clustering, mid-range connectivity, and large-scale integration. Second, the relaxation oscillators that constitute the computational primitives of the system must perform many dynamical functions with a wide variety of time constants to enable and maximally utilize information processing through transient synchronized assemblies. Third, a variety of synapses must be achievable, ranging from binary to multistable. The strength of these synapses must adjust due to network activity, as must the update frequency. These neuron and network considerations guide the designs presented in this series of papers. 

\subsection{Optical communication}
A principal challenge of differentiated computation with integrated information is communication. The core concept of the superconducting optoelectronic hardware platform is that light is excellent for this purpose. Light excels at communication for three reasons. First, light experiences no capacitance or inductance, so dense local clustering as well as long-range connections can be achieved without charge-based wiring parasitics. Second, it is possible to signal with single quanta of the electromagnetic field, thereby enabling the energy efficiency necessary for scaling. Third, light is the fastest entity in the universe. Short communication delays are ideal for maximizing the number of synchronized oscillations a neuron can participate in as well as the size of the neuronal pool participating in a synchronized oscillation. Light-speed communication therefore facilitates large networks with rich dynamics.

We have argued elsewhere \cite{shbu2017} that the capacitance and inductance of electronic interconnects is not ideal for neural computing. These limitations are ultimately due to the charge of the electron and its mass. Signals in the brain are transmitted via ionic conduction. The operating voltage of biological neurons is near 70 mV, so the energy penalty of $C V^2/2$ is significantly reduced relative to semiconducting technologies operating at 1\,V. Yet the low mobility of ions results in very low signal velocities, severely limiting the total size of biological neural systems \cite{bu2006}. Uncharged, massless particles are better suited to communication in cognitive neural systems. Light is the natural candidate for this operation. It is possible for a single optical source to fan its signals out to a very large number of recipients. This fanout can be implemented in free space, over fiber optic networks, or in dielectric waveguides at the chip and wafer scales. For large neural systems, it will be advantageous to employ all these media for signal routing. The presence of excellent waveguiding materials and a variety of light sources inclines us to utilize optical signals with 1\,\textmu m $\le \lambda \le$ 2\,\textmu m. Additionally, because the energy of a photon and its wavelength are inversely proportional, optoelectronic circuits face a power/area trade-off. Similar circuits to those presented here could be implemented with microwave circuits, but the system size would likely be cumbersome. Operation near telecommunication wavelengths appears to strike a suitable compromise. 

\subsection{Superconducting electronics}
The foundational conjecture of the proposed hardware platform is that light is optimal for communication in cognitive systems. The subsequent conjecture is that power consumption will be minimized if single photons of light can be sent and received as signals between neurons in the system. Superconducting single-photon detectors are the best candidate for receiving the photonic signals. In addition to selecting micro-scale light sources and dielectric waveguides, we choose to utilize superconduting-nanowire single-photon detectors \cite{gook2001,nata2012,liyo2013,mave2013} to receive photonic signals because of the speed \cite{yake2007}, efficiency \cite{mave2013}, and scalable fabrication \cite{buch2017} of these devices. 

Utilizing superconducting detectors contributes to energy efficiency in two ways. First, because a single photon is a quantum of the electromagnetic field, it is not possible to signal with less energy at a given wavelength. Second, because the device is superconducting, it dissipates near zero power when it is not responding to a detection event.

This choice of employing superconductors has several important ramifications. It requires that we operate at temperatures that support a superconducting ground state ($\approx$\,4\,K), so cryogenic cooling must be implemented. While cooling is an inconvenience, employment of superconducting detectors brings the opportunity to utilize the entire suite of superconducting electronic devices \cite{ti1996,vatu1998,ka1999}, including Josephson junctions and thin-film components such as current \cite{mcbe2014,mcab2016} and voltage \cite{zhto2018} amplifiers. Semiconductor light sources also benefit from low temperature \cite{doro2017}. 

We have emphasized that the charge and mass of electrons is a hindrance for communication. Yet the interactions between electrons due to their charge makes them well-suited to perform the computation and memory functions of neurons. In particular, the properties of superconducting devices and circuits make them exceptionally capable of achieving the complex dynamical systems necessary for cognition. To elucidate the specific type of dynamical devices we intend to employ, we now elaborate upon the strengths of relaxation oscillators for cognitive systems.

\subsection{Relaxation Oscillators}
As we have mentioned, a defining aspect of cognitive systems is the ability to differentiate locally to create many sub-processors, but also to integrate the information from many small regions into a cohesive system, and to repeat this architecture across many spatial and temporal scales. A network of many dynamical nodes, each with the capability of operating at many frequencies, gives rise to a vast state space. As computational primitives that can enable such a dynamical system, oscillators are ideal candidates. In particular, relaxation oscillators \cite{st2015,mist1990,soko1993,lued1997,huya2000,bu2006,gile2011,vepe1968,cacl1981} with temporal dynamics on multiple time scales \cite{soko1993} have many attractive properties for neural computing, which is likely why the brain is constructed of such devices \cite{ll1988}. We define a relaxation oscillator as an element, circuit, or system that produces rapid surges of a physical quantity or signal as the result of a cycle of accumulation and discharge. Relaxation oscillators are energy efficient in that they generally experience a long quiescent period followed by a short burst of activity. Timing between these short pulses can be precisely defined and detected \cite{bu2006}. Relaxation oscillators can operate at many frequencies \cite{huya2000} and engage with myriad dynamical interactions \cite{lued1997}. The oscillator's response is tunable \cite{huya2000}, they are resilient to noise because their signals are effectively digital \cite{stgo2005}, and they can encode information in their mean oscillation frequency as well as in higher-order timing correlations \cite{pasc1999,thde2001,sase2001,stse2007,brcl2010,haah2015}.

The relaxation oscillators we intend to employ as the computational primitives of superconducting optoelectronic networks can be as simple as integrate-and-fire neurons \cite{daab2001,geki2002} or more complex with the addition of features such as dendritic processing \cite{thde2001,sase2001,stse2007,brcl2010,haah2015} to inhibit specific sets of connections \cite{budr2004,bu2006,robu2015} or detect timing correlations and sequences of activity \cite{sase2001,haah2015}. While our choice to use superconductors was motivated by the need to detect single photons, we find superconducting circuits combining single-photon detectors and Josephson junctions are well-suited for the construction of relaxation oscillators with the properties required for neural circuits.

\subsection{Neuron overview}
\begin{figure}
	\centerline{\includegraphics[width=8.6cm]{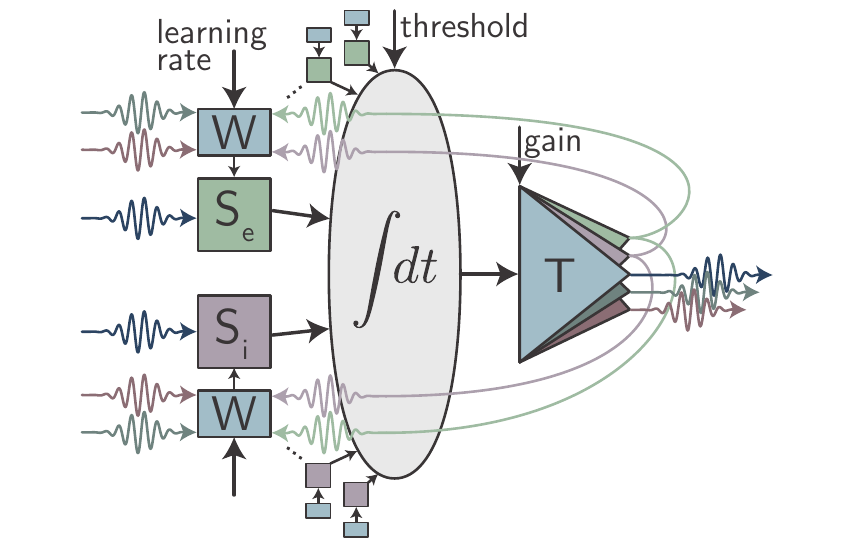}}
	\caption{\label{fig:general_schematic}Schematic of a loop neuron. Excitatory ($\mathsf{S_e}$) and inhibitory ($\mathsf{S_i}$) synapses are shown, as are the synaptic weight update circuits ($\mathsf{W}$). The wavy, colored arrows are photons, and the straight, black arrows are electrical signals. The synapses receive signals as faint as a single photon and add supercurrent to an integration loop. Upon reaching threshold, a signal is sent to the transmitter circuit ($\mathsf{T}$), which produces a photon pulse. Some photons from the pulse are sent to downstream synaptic connections, while some are used locally to update synaptic weights.}
\end{figure}
We refer to relaxation oscillators sending few-photon signals that are received with superconducting detectors as superconducting optoelectronic neurons. In the specific neurons studied in this work, integration, synaptic plasticity, and dendritic processing are implemented with inductively coupled loops of supercurrent. We therefore refer to devices of this type as loop neurons. The loop neuron presented in the remaining papers in this series is shown schematically in Fig. \ref{fig:general_schematic}. Its operation is as follows. 

Photons from afferent neurons are received by superconducting single-photon detectors at a neuron's synapses. Using Josephson circuits, these detection events are converted into an integrated supercurrent which is stored in a loop. The amount of current that gets added to the integration loop during a photon detection event is determined by the synaptic weight. The synaptic weight is dynamically adjusted by another circuit combining single-photon detectors and Josephson junctions. When the integrated current of a given neuron reaches a (dynamically variable) threshold, an amplification cascade begins, resulting in the production of light from a waveguide-integrated semiconductor light emitter. The photons thus produced fan out through a network of dielectric waveguides and arrive at the synaptic terminals of other neurons where the process repeats.

In these loop neurons, a synapse consists of a single-photon detector in parallel with a Josephson junction (which together transduce photons to supercurrent), and a superconducting loop, which stores a current proportional to the number of detected photon arrival events. This loop is referred to as the synaptic integration loop. Within each neuron, the loops of many synapses are inductively coupled to a larger superconducting loop, thereby inducing an integrated current proportional to the current in all its synapses. When the current in this neuronal integration loop reaches a threshold, the neuron produces a current pulse in the form of a flux quantum. This current is amplified and converted to voltage to produce photons from a semiconductor $p-i-n$ junction.

The currents in the synaptic and neuronal integration loops are analogous to the membrane potential of biological neurons \cite{daab2001}, and the states of flux in these loops are the principal dynamical variables of the synapses and neurons in the system. The dendritic processing functions discussed above can be implemented straightforwardly by adding intermediate mutually inductively coupled loops between the synaptic and neuronal loops. Inhibitory synapses can be achieved through mutual inductors with the opposite sign of coupling. Synapses can be grouped on dendritic loops capable of local, nonlinear processing and inhibition, analogous to dendrites \cite{sase2001,bu2006,robu2015}. Dendrites capable of detecting specific sequences of synaptic firing events \cite{thde2001,haah2015} can also be achieved. Neurons with multiple levels of dendritic hierarchy can be implemented as multiple stages of integrating loops. Clustering synapses on multiple levels of hierarchy in this way enables information access at gradually larger length scales across the network through transient synchronization at gradually lower frequencies \cite{stsa2000}. The temporal scales of the loops can be set with $L/r$ time constants, so different components can operate on different temporal scales, enabling relaxation oscillators with rich temporal dynamics. These relaxation oscillators can be combined in networks with dynamic functional connectivity, reconfigurable through inhibition \cite{robu2015,fr2015}. These receiver circuits and integration loops are presented in Ref.\,\onlinecite{sh2018b}.

Synaptic memory is also implemented based on the stored flux in a loop, referred to as the synaptic storage loop. The state of flux in the synaptic storage loop determines the current bias to the synaptic receiver circuit discussed above. This current bias is the synaptic weight. If the synaptic storage loop is created with a superconducting wire of high inductance, the loop can hold many discrete states of flux, and therefore can implement many synaptic weights. In Ref.\,\onlinecite{sh2018c} we investigate synapses with a pseudo-continuum of hundreds of stable synaptic levels between minimal and maximal saturation values, and we show that transitions between these levels can be induced based on the relative arrival times of photons from the pre-synaptic and post-synaptic neurons, thereby establishing a means for spike-timing-dependent plasticity with one photon required for each step of the memory update process. 

While synapses with many stable levels are advantageous to extending memory retention times \cite{fuab2007}, it is also important to implement synapses that change not only their efficacy based on pre- and post-synaptic spike timing, but also change their probability of changing their efficacy \cite{fudr2005}. Just as the synaptic weight is adjusted through a current bias on the receiver circuit, the probability of changing the synaptic weight can be adjusted through a current bias on the synaptic update circuit. As in the dendrites, we see a hierarchy can be achieved. In the case of synaptic memory, the synaptic weight and its rates of change are implemented in a loop hierarchy, and the state of flux in the loops can be dynamically modified based on photon detection events. Similar mechanisms can be utilized to adjust the synaptic weight based on short-term activity from the pre-synaptic neuron \cite{abre2004} or on a slowly varying temporal average of post-synaptic activity \cite{bico1982,cobe2012}. The synaptic memory circuits we develop in Ref.\,\onlinecite{sh2018c} are logical extensions of binary memory cells utilized in superconducting digital electronics \cite{vatu1998,ka1999}.

The aspect of superconducting optoelectronic neuron operation that is most difficult to achieve is the production of light. The superconducting electronic circuits that perform the aforementioned synaptic and neuronal operations operate at millivolt levels, whereas production of the telecom photons desirable for communication requires a volt across a semiconductor diode. When a neuron reaches threshold, an amplification sequence begins. Current amplification is first performed, and the resulting large supercurrent is used to induce a superconducting-to-normal phase transition in a length of wire. When the current-biased wire becomes resistive, a voltage is produced via Ohm's law. This device leverages the extreme nonlinearity of the quantum phase transition to quickly produce a large voltage and an optical pulse. The photons of this pulse are distributed over a large axonal network of passive dielectric waveguides. These waveguides terminate at each of the downstream synaptic connections. A downstream synaptic firing event will occur with near-unity probability at any connection receiving one or more photons. Photons of multiple colors can be generated simultaneously or independently, and different colors can share routing waveguides, while being used for different functions on the receiving end, such as synaptic firing and synaptic update. The number of photons produced during a neuronal firing event is the gain of the neuron, and the gain can be manipulated with the current bias to the light emitter. These transmitter circuits are discussed in Ref.\,\onlinecite{sh2018d}, and the network of waveguides that routes the communication events is discussed in Ref.\,\onlinecite{sh2018e}.

To make the analogy to biological neural hardware explicit, synapses are manifest as circuits comprising superconducting single-photon detectors with Josephson junctions. These synapses transduce photonic communication signals to supercurrent for information processing. The dendritic arbor is a spatial distribution of synapses interconnected with inductively coupled loops for intermediate integration and nonlinear processing. The integration function of the soma is also achieved with a superconducting loop, and the threshold is detected when a Josephson junction in this loop is driven above its critical current. The firing function of the soma (or axon hillock) is carried out by a chain of superconducting current and voltage amplifiers that drive a semiconductor diode to produce light. The axonal arbor is manifest as dielectric waveguides that route photonic signals to downstream synaptic connections. 

Loop neurons combine several core devices: superconducting single-photon detectors \cite{gook2001,nata2012,liyo2013,mave2013}, Josephson junctions \cite{ti1996,vatu1998,ka1999}, superconducting mutual inductors \cite{miha2005}, superconducting current \cite{mcbe2014,mcab2016} and voltage amplifiers \cite{zhto2018}, semiconductor light sources \cite{shbu2017,buch2017}, and passive dielectric waveguide routing networks \cite{chbu2017,sami2017}. While all the components of these neurons have been demonstrated independently, their combined operation in this neural circuit has not been shown. Yet the physical principles of their operation and the designs presented in this series of papers indicate the potential for loop neurons to achieve complex, large-scale neural systems. The straightforward implementation of inhibition; the realization of a variety of temporal scales through $L/r$ time constants; single-photon-induced synaptic plasticity; and dynamically variable learning rate, threshold, and gain indicate these relaxation oscillators are promising as computational primitives. In conjunction with dense local and fast distant communication over passive waveguides, the system appears capable of the spatial and temporal information integration necessary for cognition and binding. 

\subsection{The neuronal pool}
We have argued that light can achieve the connectivity necessary for information integration. There is another quantity that leads us to consider light an ideal messenger in neural systems. This quantity is the total number of neurons that can communicate with one another, referred to as the neuronal pool \cite{bu2006}. The size of the neuronal pool is treated in more detail in Ref.\,\onlinecite{sh2018e}. Here we summarize the salient result. 

If we consider networks with predominantly two-dimensional long-range connectivity (as we find in the mammalian cortex and we expect from lithographic fabrication), the number of neurons in the pool scales as the square of the signal velocity divided by the device size, $(v/w)^2$. While devices in the brain are extremely small, signal propagation is not particularly fast (2\,m/s in cortex). Optical signals are seven orders of magnitude faster than this, so even if neural systems employing optics have significantly larger devices, the size of the neuronal pool can significantly exceed what is achievable in biological systems. We estimate the neuronal pool of a superconducting optoelectronic network could comprise as many as a trillion times the number of neurons in the neuronal pool of a biological system. 

For cognition, bigger is likely better, as long as new devices represent new information, and the new information can be integrated across the system. Communication and energy efficiency are therefore principal concerns. Optical communication enables massive connectivity, and single-photon detection ensures power density never limits scaling. These considerations illustrate the potential for large-scale cognitive systems utilizing light for communication and superconductors for computation. We take an infinitesimal step toward designing networks of these neurons in Ref.\,\onlinecite{sh2018e}.
	
\section{\label{sec:discussion}Discussion}
Cognitive systems require differentiated processing and integration of information. Networks with power law spatial and temporal distributions meet these information-processing requirements. Communication is paramount both locally and globally. We conjecture that the requirement of reflecting this significance in hardware suggests we use light for communication. Micro-scale semiconducting devices are ideal light sources for dense neural integration. The requirement of power efficiency steers us to use few quanta of the electromagnetic field as our signals, a possibility enabled by superconducting detectors. This study of superconducting optoelectronic neurons combining semiconducting light sources, single-photon detectors, Josephson junctions, and dielectric waveguides indicates exceptional potential to achieve the neural functions underlying cognition. The large-scale implementation of such systems is particularly intriguing due to light-speed signals and superconductor efficiencies.

We do not propose superconducting optoelectronic networks (SOENs) as an alternative to established neural hardware, but rather as a symbiotic technology. The success of neural CMOS (including optical communication above a certain spatial scale) will contribute to the success of SOENs, as it will be advantageous for SOENs to interface with CMOS via photonic signaling on fiber optic links between cryogenic and ambient environments. SOEN hardware is particularly well suited to interfacing with other cryogenic technologies such as imaging systems with superconducting sensors  \cite{alve2015,chsc2017}, as are commonly employed for medical diagnostics \cite{hada2016}, exoplanet search \cite{raca2016,boga1992,kila2016}, cosmology \cite{diad2017}, and particle detectors \cite{le2017}. An intriguing application is in conjunction with other advanced computing technologies such as flux-based logic \cite{li2012,taoz2013,hehe2011} and quantum computers \cite{we2017}. One can envision a hybrid computational platform \cite{deli2017,posc2017} wherein a quantum computer searches the space of network weights, the neural computer learns the behavior of the quantum system, and classical fluxon logic controls the operation of both. A superconducting optoelectronic hardware platform is likely to satisfy the computation and communication requirements of this hybrid technology. 

The arguments in this paper are general, and in the subsequent four papers we present the details of the devices, circuits, and networks intended to achieve neural operation. Reference \onlinecite{sh2018b} presents the design of receiver circuits that detect photonic signals and convert them to an integrated supercurrent. We discuss the implementation of inhibition as well as dendritic processing, which are useful for dynamically tuning oscillation frequencies. The short refractory period combined with tunable response frequencies enables dynamic activity across many orders of magnitude in frequency. 

In Ref. \onlinecite{sh2018c} we introduce synaptic memory and show that it can be modified on time scales as short as 50\,ps or as long as desired. Memory update can be implemented externally for machine learning or by the internal activity of pre- and post-synaptic neurons, with each step of the memory update process requiring a single photon. We design simple, binary synapses as well as synapses with many internal plastic and metaplastic states, which achieve a balance between quick memory response and long-term recall. 

A challenge when integrating superconducting and semiconducting circuits is inducing the $\approx$\,1\,V required to drive semiconductors with low-voltage superconducting circuits. This operation is necessary if signals weighted and integrated in the superconducting domain are to produce optical signals for communication during a neuronal firing event. An amplifier circuit that produces the necessary voltage to drive the light sources is presented in Ref.\,\onlinecite{sh2018d}. A device utilizing the superconductor/metal phase transition achieves the required nonlinearity. 

In Ref.\,\onlinecite{sh2018e} we design networks of dielectric waveguides connecting semiconductor optical sources to superconducting synapses. We show that networks of a million neurons firing up to 20\,MHz, hundreds of millions of plastic synapses, and power law degree distribution can be integrated in a single complex network on a 300\,mm wafer. The power dissipated by the network would be 1\,W, a value easily managed by a standard $^4$He cryostat. We close that paper with speculation regarding the limits of neural computing in systems with light-speed communication. 

\vspace{0.5em}
This is a contribution of NIST, an agency of the US government, not subject to copyright.

\bibliography{bibliography_modelingSOENs}

\end{document}